\documentclass[runningheads]{llncs}
\usepackage{graphicx}
\usepackage{booktabs}
\usepackage{multirow}
\usepackage{siunitx}
\usepackage{grffile}
\usepackage{subcaption}
\usepackage{capt-of}
\usepackage{bm}
\usepackage{amssymb}
\usepackage{amsmath}
\usepackage{hyperref}
\usepackage{isomath}
\usepackage{wrapfig}
\usepackage{xcolor}

\usepackage{tabularx}
\usepackage{bbm}

\begin{document}
\title{FedHarmony: Unlearning Scanner Bias with Distributed Data}
%
%
\author{Nicola K. Dinsdale\inst{1} \and
Mark Jenkinson\inst{2,3,4} \and
Ana I.L. Namburete \inst{1}}
%
\authorrunning{N. K. Dinsdale  et al.}
\institute{Oxford Machine Learning in NeuroImaging (OMNI) Lab, University of Oxford, UK \and Wellcome centre for Integrative Neuroimaging, FMRIB, University of Oxford, Oxford, UK \and Australian Institute for Machine Learning (AIML), Department of Computer Science, University of Adelaide, Adelaide, Australia \and South Australian Health and Medical Research Institute (SAHMRI), North Terrace, Adelaide, Australia
\\
\email{nicola.dinsdale@cs.ox.ac.uk}}
%
\maketitle              
\begin{abstract}
The ability to combine data across scanners and studies is vital for neuroimaging, to increase both statistical power and the representation of biological variability.  However, combining datasets across sites leads to two challenges: first, an increase in undesirable non-biological variance due to scanner and acquisition differences - the \textit{harmonisation problem} - and second, data privacy concerns due to the inherently personal nature of medical imaging data, meaning that sharing them across sites may risk violation of privacy laws. To overcome these restrictions, we propose \texttt{FedHarmony}: a harmonisation framework operating in the federated learning paradigm. We show that to remove the scanner-specific effects, we only need to share the mean and standard deviation of the learned features, helping to protect individual subjects' privacy. We demonstrate our approach across a range of realistic data scenarios, using real multi-site data from the ABIDE dataset, thus showing the potential utility of our method for MRI harmonisation across studies. Our code is available at \url{https://github.com/nkdinsdale/FedHarmony}.

\end{abstract}
\section{Introduction}
\label{sec:Introduction}
Although some large scale projects, such as the UK Biobank, exist, the majority of neuroimaging datasets remain small. This necessitates the combination of data from multiple sites and scanners, both for statistical power and to represent the breadth of biological variability. However, combining data across scanners leads to an increase in non-biological variance, due to differences such as acquisition protocol and hardware \cite{Han2006,Jovicich2006,Yu2018}. Thus, we need \textit{harmonisation methods} to enable joint unbiased analysis of data from different scanners and studies. 

However, data cannot simply be pooled (combined) across imaging sites without ensuring compliance with data privacy laws, particularly if we wish to use representative clinical imaging data, as the sharing of medical images is covered by legislation such as GDPR \cite{GDPR} and HIPAA \cite{HIPPA}. Federated learning (FL) has been proposed as a method to train models on distributed data \cite{McMahan2017CommunicationEfficientLO}, where data are kept on their local servers, the users train local models on their own private datasets and then share only the weights or gradients of the trained models. FL has the potential to become the standard paradigm for multisite imaging, with early studies demonstrating its feasibility in neuroimaging  \cite{Sheller2019}.  

Direct translation of existing harmonisation methods into FL frameworks is non-trivial. Most deep learning methods for harmonisation are based on generative frameworks \cite{Dewey2019,Moyer2020,Zhao2019,Zuo2021}, and, although federated equivalents to GANs and VAEs are being developed \cite{Rasouli2020,Zhang2021}, additional challenges exist for harmonisation approaches that require simultaneous access to source and target data \cite{Feng2021KD3AUM}. Additionally, many methods require paired data  -- not possible with distributed data and  unlikely to exist in large multisite studies. Further, most generative methods are data-hungry \cite{Dewey2019,Zuo2021}, which casts into doubt whether sufficient data would be available at local sites. 

Alternative methods for harmonisation frame the task as a domain adaptation problem (DA) \cite{Dinsdale2021b}, as the goal mirrors the harmonisation problem: removal of information regarding domain (scanner or acquisition protocol) while retaining the true variance of interest (the biological signal). In \cite{Dinsdale2021b}, the DA approach has been demonstrated for a range of tasks and network architectures. The harmonisation approach in \cite{Dinsdale2021b} is built upon the DA framework proposed in \cite{Tzeng2015SimultaneousDT} and has since been translated into the federated setting \cite{Peng2020}. In \cite{Peng2020} the domain adaptation across local sites is achieved by sharing feature embeddings for local data points in a global shared knowledge bank, such that a global domain predictor can be trained. However, sharing the features still leads to privacy concerns \cite{Feng2021KD3AUM}, as images are potentially recoverable from the features. Many of the other FL DA methods proposed cannot be adapted for the harmonisation problem, as, for instance, they produce domain-specific models or ensembles \cite{Feng2021KD3AUM,Peterson2019PrivateFL,Yang2020,Yao2021}, where the final predictions depend on the source of the data.

Therefore, we present \texttt{FedHarmony}: a method to adapt the harmonisation framework proposed in \cite{Dinsdale2021b} for distributed data, while minimising shared information. We aim to train a model that performs equally well on the task of interest across all of the imaging sites while using features that are invariant to the imaging site, through a horizontal federated framework \cite{Yang2019}. We show that by approximating the learned feature embeddings as Gaussian distributions, we can share only a mean and standard deviation per feature which, by definition, contain no identifying information for individual subjects. Alongside standard privacy-protecting approaches \cite{Dwork2006,Sweeney2002}, sharing only these summary statistics would make the approach robust to honest-but-curious adversaries \cite{Yang2019}. By demonstrating our approach, \texttt{FedHarmony}, for a range of data settings, we show its viability for data harmonisation for multisite distributed imaging studies. 

\section{Methods}
\label{sec:Methods}
\begin{figure}
\centering
\includegraphics[width=0.85\textwidth]{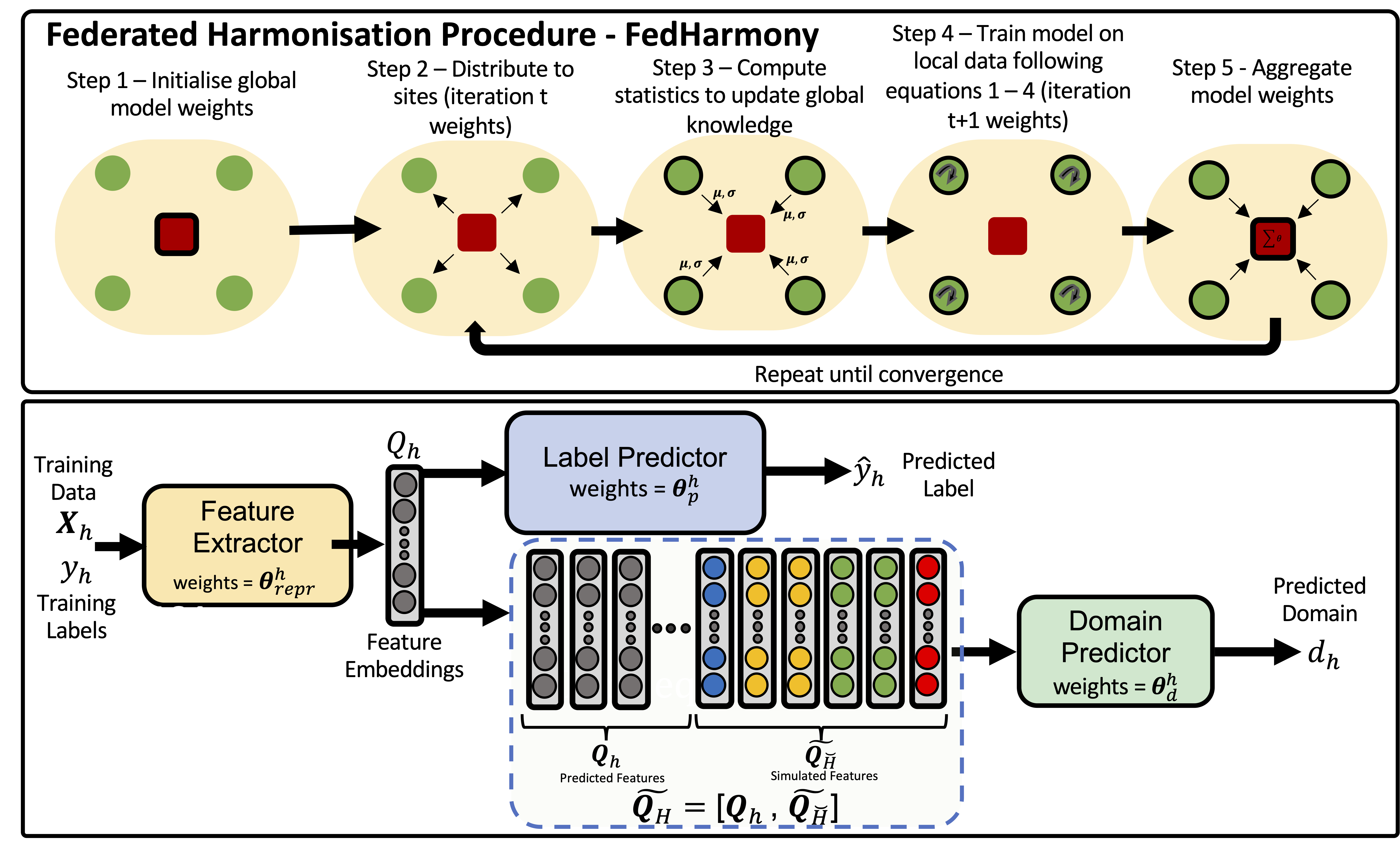}
\caption{\texttt{FedHarmony} Procedure and high level general network architecture.}
\label{fig:architecture}
\end{figure}
\noindent Consider the scenario in which we have multisite MRI data, where the data for each site $h$ are stored on their local server, for all sites $H$. For each site, we have pairs of training data $D_h = \{\bm{X}_h, \bm{y}_h\}$ where $\bm{X}_h$ is the input data and $\bm{y}_h$ is the label, and we consider each site to be a separate data domain. We wish to create a model that performs as well as possible on each imaging site, whilst having predictions that are invariant to the acquisition site. A global model is constructed, which following \cite{Dinsdale2021b} is formed of a feature extractor (with weights $\bm{\theta}_{repr}$), a label predictor ($\bm{\theta}_{p}$) and a domain predictor ($\bm{\theta}_{d}$) (Fig. \ref{fig:architecture}). This model is initialised randomly (step 1), and then the \texttt{FedHarmony} procedure begins, as shown in Fig. \ref{fig:architecture}, with the global model weights for the three network components,  $\bm{\theta}_{repr,t}$, $\bm{\theta}_{p,t}$ and $\bm{\theta}_{d,t}$, being sent to the local sites; $t$ indicates these are the current weights for iteration $t$ and thus each site receives the same initialisation (step 2). 
\\

\noindent 
\textbf{Update Global Knowledge Store (Step 3):} To remove scanner information adversarially, the different sites' nodes need to have an understanding of the various scanner characteristics. Although there is precedent for achieving this through sharing the feature embeddings from local sites to the global site \cite{Peng2020}, we aim to minimise the quantity of shared information to help protect individual privacy. It has been shown that domain-specific characteristics can be encoded to enable contrastive learning  \cite{Dong2001}, by sharing the mean and standard deviation of the feature embeddings following a BoxCox transformation \cite{BoxCox} (an invertible transform which aims to make the distribution maximally Gaussian), suggesting that this might be sufficient information to also remove scanner information. However, in many multisite MRI studies, small amounts of local data are available that may be insufficient to fit a BoxCox transformation. Thus, we simplify the transformation by assuming that the features will already be normally distributed and that we can characterise $\bm{Q}_h= featureExtractor(\bm{X}_h, \bm{\theta}_{repr,t})$ by their mean and standard deviations. Therefore the total information shared per site is a mean and standard deviation per feature, which inherently has no information about individuals, and we collate a global information store $\{ \bm{\mu}_h, \bm{\sigma}_h :h \in H\}$.

\noindent
\textbf{Training Procedure (Step 4):}
Local training is then controlled by three loss functions, based on \cite{Dinsdale2021b}. First is the main task loss $L_p$, in which the feature extractor and the label predictor are updated. If $\Theta_t=[\bm{\theta}_{repr, t}, \bm{\theta}_{p, t}]$ are the current global weights (iteration $t$) for the feature extractor and the label predictor, and $\Theta_{t+1}^h=[\bm{\theta}_{repr, t+1}^h, \bm{\theta}_{p, t+1}^h]$ are the updated ($t+1$) weights for site $h$, then:
\begin{equation}
	L_p(\bm{X}_h, \bm{y}_h; \bm{\theta}_{repr, t+1}^h, \bm{\theta}_{p, t+1}^h) = 
	\frac{1}{N_h} \sum_{j=1}^{N_h} L_{task}(\bm{y}_{h,j}, \hat{\bm{y}}_{h,j}) 
	+ \mu L_{prox}(\Theta_t, \Theta_{t+1}^h)
\end{equation}
where $L_{task}$ is the task-specific loss function, averaged over the training samples available for the given site, $N_h$, and $\bm{y}_{h,j}$ and $\hat{\bm{y}}_{h,j}$ are the true and predicted labels for the $j^{th}$ example for the $h^{th}$ imaging site. $L_{prox}$ is a proximal loss term based on \texttt{FedProx} \cite{FedProx} that penalises weight deviations from the global model that do not sufficiently improve the main task loss and has been shown to aid FL when the data is non-iid  \cite{FedProx}, as is inherently true for multisite MRI data, hence the harmonisation problem. Thus, if $\mu$ is a constant, weighting the contributions from the two losses, then: 
\begin{equation}
L_{prox}(\Theta_t, \Theta_{t+1}^h) = ||\bm{\theta}_{repr, t} - \bm{\theta}_{repr, t+1}^h || ^ 2 + || \bm{\theta}_{p, t} -  \bm{\theta}_{p, t+1}^h||^2
\label{eq:proxloss}
\end{equation}

The second loss function, $L_d$, then updates the domain predictor to be able to discriminate between the sites. Data is only available for the local site $h$ and thus can only directly generate the features, $\bm{Q}_h$, needed to train the domain predictor for this local site. For all the other sites $\breve{H}$ we generate example features randomly using the shared knowledge store; thus, for site $\breve{h}$ in $\breve{H}:\bm{\widetilde{Q}}_{\breve{h}} \sim \mathcal{N}(\bm{\mu}_{\breve{h}}, \bm{\sigma}_{\breve{h}})$. The generated features,
$\bm{\widetilde{Q}}_{\breve{H}} = \{\bm{\widetilde{Q}}_{\breve{h}} \forall \breve{h} \in \breve{H} \}$ are then concatenated with the true features, $\bm{Q}_h$, such that $\bm{\widetilde{Q}}_H^h = [\bm{Q}_h, \bm{\widetilde{Q}}_{\breve{H}}]$. The number of simulated subjects generated for each site $\breve{h}$ is chosen randomly such that the final batchsize is twice that of the original; the features and corresponding domain (site) labels are then shuffled. We then update the parameters of the domain predictor, using categorial cross-entropy to assess the site information remaining in $\bm{Q}_h$: 
\begin{equation}
 L_d (\bm{\widetilde{Q}}^h_{H}, \bm{d} ; \bm{\theta}_{d, t+1}^h) = - \sum^H_{h=1} \mathbbm{1}[d = h] log(p_h)
 \label{eq:domain}
\end{equation}
where $p_h$ are the softmax outputs of the domain classifier. The third loss is the confusion loss $L_{conf}$, which removes the site information, by penalising deviation in $p_h$ from a uniform distribution:
\begin{equation}
L_{conf}(\bm{X}_h, \bm{\theta}_{d, t+1}^h; \bm{\theta}_{repr, t+1}^h) = - \sum^H_{h=1} \frac{1}{H} log(p_h)
\label{eq:confusion}
\end{equation}
Therefore, the overall method can be considered to minimise the total loss function $L = L_p + \alpha L_d + \beta L_{conf}$ where $\alpha$ and $\beta$ control the relative contributions of the different loss functions. The domain loss (Eq. \ref{eq:domain}) and the confusion loss (Eq. \ref{eq:confusion}) act in opposition to each other, and so must be updated iteratively, with three iterations per epoch. The procedure is iterated through for $E$ local epochs.
\\

\noindent \textbf{Weight Aggregation (Step 5):} After each site has completed the $E$ local epochs, the local weights are returned to the global node to be aggregated. In \cite{Dinsdale2021b} it was shown that the harmonisation process was aided through evaluating each site separately in the loss function rather than averaging over all data points. Given that \texttt{FedAvg} \cite{McMahan2017CommunicationEfficientLO} is equivalent to averaging over all training subjects for FL, we alter the aggregation to have equal contributions from each site: 
$ \bm{\theta}_{t+1} \leftarrow  \frac{1}{H} \sum^H_{h=1} \bm{\theta}_{t+1}^{h}$
where the aggregation is completed separately for $\bm{\theta}_{repr}$, $\bm{\theta}_{p}$ and $\bm{\theta}_{d}$. We term this aggregation step \texttt{FedEqual}. Once the aggregation is complete, the training loops through stages 2 - 5 (Fig. \ref{fig:architecture}) until convergence. 
\\

\noindent
\textbf{Semi-supervised setting:}
We then explore the scenario where labels are only available for 
a subset of sites, which is realistic, as labels are expensive to generate in terms of time and expertise. Thus, we also consider the semi-supervised setting, adjusting the framework such that local training is only completed for sites with labels available, but we continue to update the global knowledge store for all sites. Therefore, the removal of the site information still considers all sites. 

\section{Implementation Details}
\begin{figure}
\centering
\includegraphics[width=0.5\textwidth]{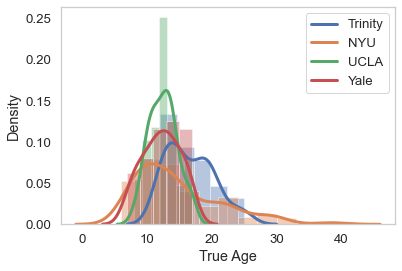}
\caption{Normalised age distributions for the 4 sites from the ABIDE dataset: Trinity: 49 subjects, 16.7$\pm$3.6 years (range: 12-25), NYU: 182 subjects, 14.7$\pm$6.6 (6-39), UCLA: 99 subjects, 12.5$\pm$2.2 (8-17), Yale: 56 subjects, 12.2$\pm$2.8 (7-17).}
\label{fig:age_dist}
\end{figure}

For our experiments we use data from the ABIDE dataset\footnote{Data from: \url{https://fcon_1000.projects.nitrc.org/indi/abide/}} \cite{ABIDE}, for the task of age prediction, using T1 MR images as $\bm{X}$ and the ages as the labels $\textbf{y}$. The MR images from each site were processed using \texttt{FSL anat}\footnote{\url{https://fsl.fmrib.ox.ac.uk/fsl/fslwiki/fsl_anat}}. Subjects were rejected where the pipeline failed or where data was missing. Four sites (Trinity, NYU, UCLA, Yale) were chosen for our experiments, so as to span both age distributions and subject numbers (Fig. \ref{fig:age_dist}). The data were split into training/validation/test sets as $70\%/10\%/20\%$, yielding a maximum of 127 subjects for training (NYU) and a minimum of 35 (Trinity). Our networks and baselines were all trained with 5-fold cross-validation until no improvement in the average mean absolute error (MAE) was reported on the validation data for all local sites. We present the results on the held-out testing data in all cases.

For the network, we used the VGG-based architecture used in \cite{Dinsdale2021b}\footnote{code from: \url{https:/github.com/nkdinsdale/Unlearning_for_MRI_harmonisation}}. The images were resized to (128, 240, 160) and normalised to have zero mean and unit standard deviation. As demonstrated by \cite{Dinsdale2021b}, the framework is flexible and should be applicable to many feedforward architectures and tasks, with age prediction being used to demonstrate the approach due to the availability of labels across many sites. The implementation was in Python 3.6.8 and PyTorch 1.10.2. The networks were trained on a 16GB P100 GPU, taking up to 5 minutes per local epoch (NYU), with a batchsize of 16, and requiring 50-75 rounds of communication to converge (\texttt{FedHarmomy} required 61 rounds and \texttt{FedAvg} required 56). The following hyperparameters were used for all experiments based on \cite{Dinsdale2021b,FedProx}, with the same values applied for each local site: batchsize=16, learning rate=$1\times10^{-4}$ with Adam optimiser, local epochs (E)=10, $\mu$=0.01, $\alpha$=1, and $\beta$=100.

\section{Results \& Discussion}

\noindent\textbf{Fully Supervised:} We first compared our method to several baselines: training on NYU data only (the largest site, representing where data cannot be shared and we have no federated learning framework), training on all sites normally (centralised data), \texttt{FedAvg} \cite{McMahan2017CommunicationEfficientLO}, \texttt{FedProx} \cite{FedProx} and our proposed \texttt{FedEqual} aggregation method. We also completed an ablation study, exploring the effect of the aggregation and proximal losses alongside the harmonisation. To compare methods, we evaluated both the mean absolute error (MAE), where we want the best performance possible across all sites, and the scanner classification accuracy (SCA): the performance of a domain classifier trained on $\bm{Q}$ at convergence, where the aim is to achieve random chance (25\%). Note that the domain predictor was able to achieve 95\% on the feature embeddings when the age prediction was trained with NYU only, and so is clearly able to identify scanner information.  

\begin{figure}
     \centering
     \begin{subfigure}[b]{0.4\textwidth}
         \centering
         \includegraphics[width=\textwidth]{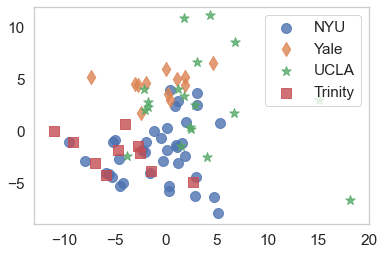}
         \caption{\texttt{FedAvg}}
     \end{subfigure}
     \begin{subfigure}[b]{0.4\textwidth}
         \centering
         \includegraphics[width=\textwidth]{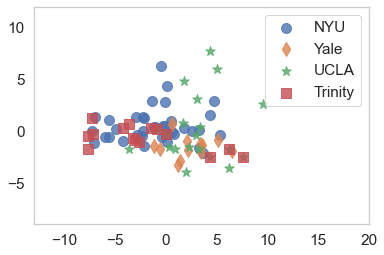}
         \caption{\texttt{FedHarmony}}
     \end{subfigure}
        \caption{PCA of $\bm{Q}$ for \texttt{FedAvg} and \texttt{FedHarmony}, showing, as expected, increased overlap across sites with \texttt{FedHarmony}.}
        \label{Fig: features}
\end{figure}

Considering the results in Table \ref{tab:fullysupervised}, we were able train federated models (\texttt{FedAvg}) to perform as well as standard centralised training. \texttt{FedHarmony} led to improvement both in terms of average performance across all sites and in the reduction of SCA to close to random chance. This improvement indicates the success of the DA approach, as information from across sites is being used to aid predictions even though only summary statistics of the features are being shared. The ablation study shows the need for the three additions to the \texttt{FedAvg} baseline, with the $L_{prox}$ (Eq. \ref{eq:proxloss}) being vital for both stability of training and the ability to remove scanner information. Fig. \ref{Fig: features} shows a PCA of $\bm{Q}$ comparing \texttt{FedAvg} and \texttt{FedHarmony}, where it is evident that the harmonisation increases the feature overlap of the different sites.
\\

\begin{table}[]
\caption{\textbf{Fully supervised results}: Component represents additions relative to \texttt{FedAvg} \cite{McMahan2017CommunicationEfficientLO}: Equal = Aggregation as \texttt{FedEqual}; Prox = Proximal Loss as Eq. \ref{eq:proxloss}; Harm = Harmonisation, otherwise local training only considers the main task loss. SCA = Scanner Classification Accuracy of domain classifier retrained on $\bm{Q}$ at the end of training, where random chance (25\%) is the goal. NYU only and all sites = centralised training. \texttt{FedHarmony} is our proposed approach. * = significant improvement over next performing method (paired t-test, $p < 0.001$).}
\label{tab:fullysupervised}
\resizebox{\textwidth}{!}{%
\begin{tabular}{|l|ccc|cccc|c|c|}
\hline
\multirow{2}{*}{Method} & \multicolumn{3}{c|}{Component}                                                 & \multicolumn{4}{c|}{Site MAE}                                                              & \multirow{2}{*}{\begin{tabular}[c]{@{}c@{}}Average\\ MAE\end{tabular}} & \multicolumn{1}{c|}{\multirow{2}{*}{\begin{tabular}[c]{@{}c@{}}SCA (\%)\end{tabular}}} \\ \cline{2-8}
                        & \multicolumn{1}{c|}{Equal} & \multicolumn{1}{c|}{Prox} & Harm                    & \multicolumn{1}{c|}{NYU n=40} & \multicolumn{1}{c|}{Yale n=14} & \multicolumn{1}{c|}{UCLA n=20} & Trinity n=10 &                                                                        & \multicolumn{1}{c|}{}                                                                                              \\ \hline
NYU Only                & \multicolumn{1}{l|}{}      & \multicolumn{1}{l|}{}     & \multicolumn{1}{l|}{} & \multicolumn{1}{c|}{5.26$\pm$4.37}        & \multicolumn{1}{c|}{2.38$\pm$1.42}    & \multicolumn{1}{c|}{3.22$\pm$2.26}     & 9.66$\pm$7.89      &                                                                       5.13 & 95                                                                                                                    \\ \hline
All Sites               & \multicolumn{1}{c|}{}      & \multicolumn{1}{c|}{}     & \multicolumn{1}{c|}{} & \multicolumn{1}{c|}{5.10$\pm$4.56}        & \multicolumn{1}{c|}{2.32$\pm$1.50}    & \multicolumn{1}{c|}{2.87$\pm$2.13}     & 4.53$\pm$3.49     & 3.70                                                                        &                                                                                                                    86 \\ \hline \hline
\texttt{FedAvg \cite{McMahan2017CommunicationEfficientLO}}                  & \multicolumn{1}{c|}{}      & \multicolumn{1}{c|}{}     & \multicolumn{1}{c|}{} & \multicolumn{1}{c|}{5.26$\pm$4.74}        & \multicolumn{1}{c|}{$\bm{1.99\pm1.21}^*$}    & \multicolumn{1}{c|}{2.57$\pm$1.88}     & 4.94$\pm$3.80      &                                                                        3.69                                                                                                                  & 64                                                                                                                    \\ \hline
\texttt{FedProx \cite{FedProx}}                   & \multicolumn{1}{c|}{}      & \multicolumn{1}{c|}{\checkmark}    &                       & \multicolumn{1}{c|}{5.21$\pm$4.70}        & \multicolumn{1}{c|}{2.15$\pm$1.44}    & \multicolumn{1}{c|}{2.61$\pm$1.73}     & 4.47$\pm$3.56      &                                                                       3.61 & 62                                                                                                                    \\ \hline
\texttt{FedEqual}                 & \multicolumn{1}{c|}{\checkmark}     & \multicolumn{1}{c|}{}     &                       & \multicolumn{1}{c|}{6.07$\pm$3.99}        & \multicolumn{1}{c|}{2.06$\pm$1.31}    & \multicolumn{1}{c|}{2.61$\pm$2.30}     & 5.43$\pm$3.00      & 4.06 & 42 \\ \hline \hline
Ablation A              & \multicolumn{1}{c|}{}      & \multicolumn{1}{c|}{}     & \checkmark                     & \multicolumn{1}{c|}{12.21$\pm$10.31}        & \multicolumn{1}{c|}{4.20$\pm$3.61}    & \multicolumn{1}{c|}{6.21$\pm$3.21}     & 10.32$\pm$8.31     & 8.23                                                                        &                                                                                                                   95 \\ \hline
Ablation B               & \multicolumn{1}{c|}{\checkmark}     & \multicolumn{1}{c|}{}     & \checkmark                     & \multicolumn{1}{c|}{5.27$\pm$4.90}        & \multicolumn{1}{c|}{2.27$\pm$1.13}    & \multicolumn{1}{c|}{2.66$\pm$1.98}     &  4.81$\pm$3.98   &                                                                        3.75 &                                                                                                                   34 
 \\ \hline
Ablation C            & \multicolumn{1}{c|}{}      & \multicolumn{1}{c|}{\checkmark}    & \checkmark                     & \multicolumn{1}{c|}{\bm{$4.93\pm4.31}^*$}        & \multicolumn{1}{c|}{2.13$\pm$1.91}    & \multicolumn{1}{c|}{2.26$\pm$1.15}     & 4.36$\pm$3.37       &                                                                       $\bm{3.42}$ & 35 \\ \hline \hline
\texttt{FedHarmony}              & \multicolumn{1}{c|}{\checkmark}     & \multicolumn{1}{c|}{\checkmark}    & \checkmark                     & \multicolumn{1}{c|}{5.14$\pm$4.52}        & \multicolumn{1}{c|}{2.04$\pm$1.25}    & \multicolumn{1}{c|}{$\bm{2.13\pm1.45}^*$}     &  $\bm{4.34\pm3.65}$   &                                                                       $\bm{3.42}$ &                                                                                                                   $\bm{29}$ \\ \hline
\end{tabular}%
}
\end{table}

\begin{table}[]
\centering
\caption{\textbf{Semi-supervised results}: \textit{NYU only} = centralised training only NYU; \textit{No Trinity} = centralised training no Trinity; \textit{All sites} = centralised training. SCA = Scanner Classification Accuracy of domain classifier retrained on $\bm{Q}$ at the end of training, where random chance (25\%). \texttt{FedHarmony} was trained locally for only labelled sites, and global shared information used for all sites. * = significant improvement over next performing method (paired t-test, $p < 0.001$).}
\label{tab:semisupervised}
\resizebox{0.8\textwidth}{!}{%
\begin{tabular}{|l|cccc|c|c|}
\hline
\multirow{2}{*}{Method} & \multicolumn{4}{c|}{Site MAE}                                                                                   & \multicolumn{1}{l|}{\multirow{2}{*}{\begin{tabular}[c]{@{}c@{}}Average\\ MAE\end{tabular}}} & \multirow{2}{*}{\begin{tabular}[c]{@{}c@{}}SCA (\%) \end{tabular}} \\ \cline{2-5}
                        & \multicolumn{1}{c|}{NYU n=40} & \multicolumn{1}{c|}{Yale n=14} & \multicolumn{1}{c|}{UCLA n=20} & \multicolumn{1}{c|}{Trinity n=10} & \multicolumn{1}{c|}{}                                                                       &                                                                                              \\ \hline \hline
 \multicolumn{7}{|c|}{a) Semi-supervised 1 Site} \\                           
                         \hline
NYU Only                & \multicolumn{1}{c|}{$\bm{5.26\pm4.37}$}        & \multicolumn{1}{c|}{3.22$\pm$2.26}    & \multicolumn{1}{c|}{3.22$\pm$2.26}     &     9.66$\pm$7.89                      &                                                                                            5.13 &                                                                                              95 \\ \hline
\texttt{FedHarmony}              & \multicolumn{1}{c|}{$\bm{5.26\pm4.15}$}        & \multicolumn{1}{c|}{$\bm{2.20\pm1.79}^*$}    & \multicolumn{1}{c|}{$\bm{2.91\pm1.56}^*$}     &  $\bm{4.09\pm3.32^*}$                         &           \textbf{3.61$^*$} &                                                                                              \textbf{30} \\ \hline \hline
 \multicolumn{7}{|c|}{b) Semi-supervised 3 Site} \\                           \hline
 No Trinity              & \multicolumn{1}{c|}{5.36$\pm$4.65}    & \multicolumn{1}{c|}{$\bm{1.81\pm1.66}^*$}     & \multicolumn{1}{c|}{$\bm{1.83\pm1.16}^*$}     &    7.29$\pm$5.00                          &          4.07                                                              &                          96 \\ \hline
\texttt{FedProx} \cite{FedProx}                & \multicolumn{1}{c|}{5.15$\pm$4.76}    & \multicolumn{1}{c|}{2.15$\pm$1.38}     & \multicolumn{1}{c|}{2.62$\pm$1.82}     &                             5.96$\pm$4.05 &                                                                       3.96 &    74                       \\ \hline
\texttt{FedHarmony}              & \multicolumn{1}{c|}{$\bm{5.12\pm4.60}^*$}    & \multicolumn{1}{c|}{2.39$\pm$1.67}     & \multicolumn{1}{c|}{2.16$\pm$1.40}     &                             $\bm{4.69\pm4.19}^*$ &                                                                       \textbf{3.59$^*$} & \textbf{26 }                         \\ \hline \hline

All Sites               & \multicolumn{1}{c|}{5.10$\pm$4.56}        & \multicolumn{1}{c|}{2.32$\pm$1.50}    & \multicolumn{1}{c|}{2.87$\pm$2.13}     & 4.53$\pm$3.49                         &                                                                                            3.70 &                                                                                              86 \\ \hline
\end{tabular}%
}
\end{table}

\noindent \textbf{Semi-supervised:} We then simulated two semi-supervised scenarios, where training labels were only available for a subset of the sites. First we considered when only one site had available training labels, NYU, as a multitarget DA task, which could for example represent when the organising site has labelled data. In Table \ref{tab:semisupervised} we compare to standard training (\texttt{FedAvg} and \texttt{FedProx} are equivalent to standard training for a single labelled site). Using \texttt{FedHarmony}, we were able to train the network to perform as well as fully supervised centralised training, while removing the scanner information, showing the power of the shared global knowledge. We also considered when three sites have labels and one does not (Trinity), for example, if a new site joins the study. \texttt{FedHarmony} led to improvement in performance again, even compared to having labels for all sites, while removing the scanner information. These results show the suitability of our approach across realistic data scenarios for multisite MRI studies.
\\

\begin{figure}
\centering
\includegraphics[width=0.8\textwidth]{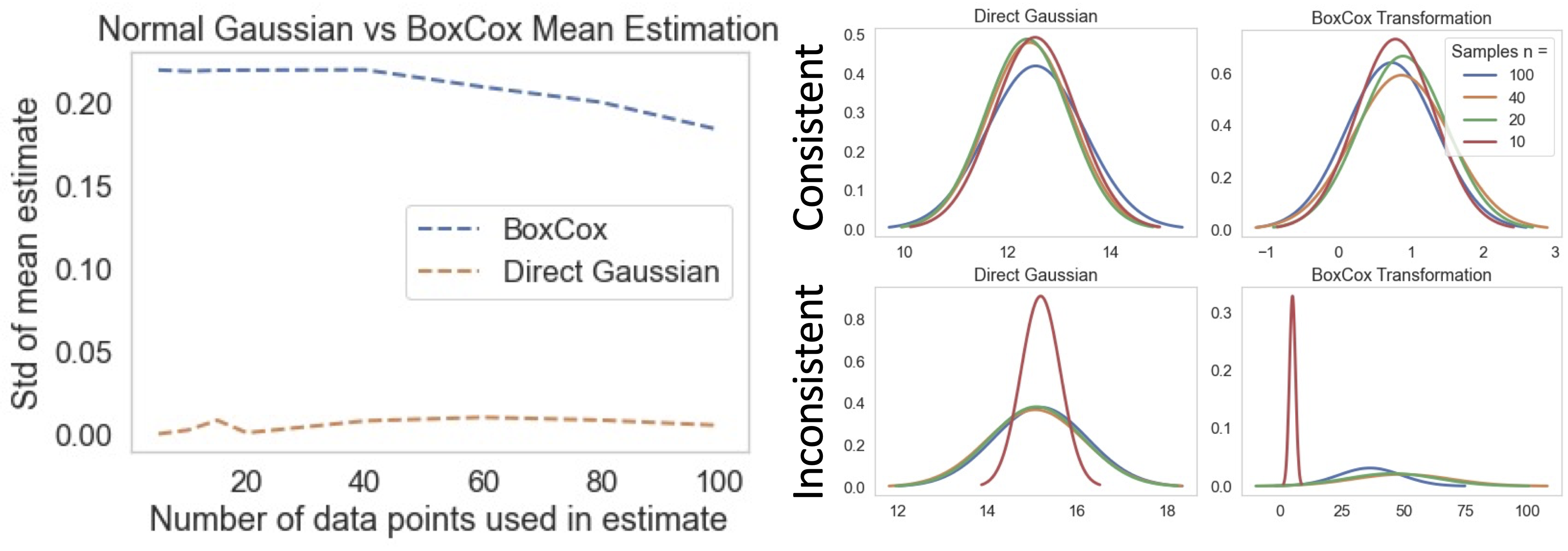}
\caption{We plot the standard deviation of 100 mean estimates of the fit for $\bm{Q}_{NYU}$, for increasing numbers of random samples from the full $\bm{Q}_{NYU}$, alongside consistent and inconsistent fits. Note: the estimated means will differ, as the BoxCox transforms the data to be maximally Gaussian rather than fitting the data.} 
\label{fig:gaussian}
\end{figure}

\noindent \textbf{Suitability of Gaussian Fit:} In Fig. \ref{fig:gaussian} we plot the standard deviation of 100 mean estimates for $\bm{Q}_{NYU}$ from both the direct Gaussian fit and the BoxCox transformation for increasing numbers of samples. It is evident that the direct Gaussian fit leads to more consistent estimates, indicating increased suitability in low data regimes compared to the BoxCox transform, as the representation on the training data is more likely to also represent the testing data. 

\section{Conclusion}
We have presented \texttt{FedHarmony}, a method to allow harmonisation of MRI data in a federated learning scenario, through development of a domain adaptation approach with minimal information sharing, outperforming baseline and FL approaches across realistic data scenarios. Alongside standard privacy-protecting approaches, such as differential privacy, our approach would enable training models on multisite MRI data while maintaining individual subjects' privacy. 

\section{Acknowledgements}
ND is supported by a Academy of Medical Sciences Springboard Award. MJ is supported by the National Institute for Health Research, Oxford Biomedical Research Centre, and this research was funded by the Wellcome Trust [215573/Z/19/Z]. WIN is supported by core funding from the Wellcome Trust [203139/Z/16/Z]. AN is grateful for support from the UK Royal Academy of Engineering under the Engineering for Development Research Fellowships scheme and the Academy of Medical Sciences. The computational aspects of this research were supported by the Wellcome Trust Core Award [Grant Number 203141/Z/16/Z] and the NIHR Oxford BRC. The views expressed are those of the author(s) and not necessarily those of the NHS, the NIHR or the Department of Health.

%
%
 \bibliographystyle{splncs04}
\bibliography{references}
\end{document}